\pdfoutput=1

\documentclass[11pt]{article}

\usepackage[preprint]{acl}

\usepackage{times}
\usepackage{latexsym}

\usepackage[T1]{fontenc}

\usepackage[utf8]{inputenc}

\usepackage{microtype}

\usepackage{inconsolata}

\usepackage{xspace}
\usepackage{inconsolata}
\usepackage{booktabs}
\usepackage{comment}
\usepackage{graphicx} 
\usepackage{tcolorbox} 
\usepackage{framed}
\usepackage{CJKutf8} 
\usepackage{multirow}
\usepackage{url}
\usepackage{svg}
\usepackage{soul}
\usepackage{fontawesome5}
\usepackage{xltabular} 

\usepackage{amsthm}

\usepackage{tikz}
\usepackage{forest}
\usetikzlibrary{trees,positioning,shapes,shadows,arrows.meta}
\usepackage{amssymb}
\usepackage{adjustbox}
\usepackage{rotating}
\usepackage{array}
\usepackage{makecell}
\usepackage{enumitem}
\usepackage{todonotes}

\usepackage{amsmath}

\title{Multimodal Emotion Recognition in Conversations: \\A Survey of Methods, Trends, Challenges and Prospects}

\author{
  \textbf{Chengyan Wu\textsuperscript{$\ast$1,2}}, 
\textbf{Yiqiang Cai\textsuperscript{$\ast$1,2}}, 
\textbf{Yang Liu\textsuperscript{3}}, 
\textbf{Pengxu Zhu\textsuperscript{4}}, 
\\
\textbf{Yun Xue\textsuperscript{$\ddagger$1,2}}, 
\textbf{Ziwei Gong\textsuperscript{5}}, 
\textbf{Julia Hirschberg\textsuperscript{5}}, 
\textbf{Bolei Ma\textsuperscript{$\ddagger$6}}
\vspace{4.5pt}
\\
\small{\textsuperscript{1}Guangdong Provincial Key Laboratory of Quantum Engineering and Quantum Materials}\\
\small{\textsuperscript{2}School of Electronic Science and Engineering (School of Microelectronics), South China Normal University}\\
\small{\textsuperscript{3}North Carolina Central University}~
\small{\textsuperscript{4}Georgia Institute of Technology}~
\small{\textsuperscript{5}Columbia University}~\\\small{\textsuperscript{6}LMU Munich \& Munich Center for Machine Learning}
\vspace{3.5pt}
\\
\small{$^\ast$Equal contributions. $^\ddagger$Corresponding authors.}
\vspace{3pt}
\\
\small{
\texttt{\{chengyan.wu, yiqiangcai, xueyun\}@m.scnu.edu.cn, zg2272@columbia.edu, bolei.ma@lmu.de}
}
}

\begin{document}
\maketitle
\begin{abstract}
While text-based emotion recognition methods have achieved notable success, real-world dialogue systems often demand a more nuanced emotional understanding than any single modality can offer. Multimodal Emotion Recognition in Conversations (MERC) has thus emerged as a crucial direction for enhancing the naturalness and emotional understanding of human-computer interaction. Its goal is to accurately recognize emotions by integrating information from various modalities such as text, speech, and visual signals. 

This survey offers a systematic overview of MERC, including its motivations, core tasks, representative methods, and evaluation strategies. We further examine recent trends, highlight key challenges, and outline future directions. As interest in emotionally intelligent systems grows, this survey provides timely guidance for advancing MERC research.

\end{abstract}

\section{Introduction}
Emotion recognition in conversations (ERC) \cite{peng2022survey,deng2021survey} is an increasingly important task in natural language processing (NLP), focusing on identifying the emotional state associated with each utterance in a dialogue. Unlike conventional emotion classification on isolated sentences, ERC requires understanding the interplay between utterances and tracking speaker-specific context across the conversation \cite{DBLP:conf/mm/GaoH00SX24}. Its relevance has grown due to its potential in various real-world applications, including social media monitoring \cite{Kumar_Dogra_Dabas_2015}, intelligent healthcare services \cite{hu-etal-2021-mmgcn}, and the design of emotionally aware dialogue agents \cite{jiao2020real, gong-etal-2023-eliciting}.

\begin{figure}[htbp]
\centering
  \includegraphics[width=0.95\columnwidth]{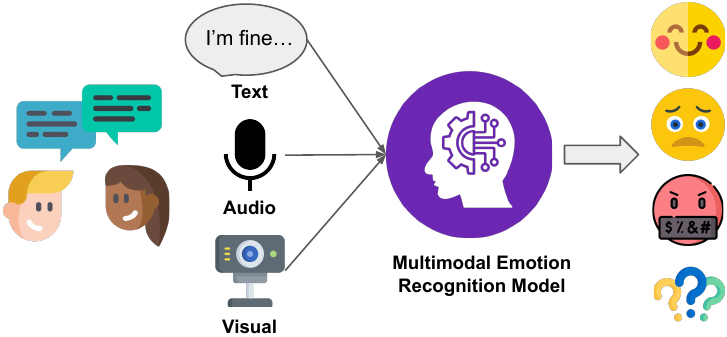}
  \caption{An example of MERC. Text, audio, and visual inputs are integrated through a multimodal model to detect various emotional states.}
  \label{tag:introduction}
\end{figure}

However, human emotions are typically conveyed through multiple modalities, including auditory (e.g., speech), visual (e.g., facial expressions, gestures), and linguistic (e.g., the semantic content conveyed by transcribed text). 
As a result, recent research \citep[e.g.,][]{ma2023transformer,van2025effective,dutta2025llmsupervisedpretrainingmultimodal} has increasingly focused on multimodal settings in dialogue, a direction we refer to as Multimodal Emotion Recognition in Conversations (MERC). Researchers aim to identify the emotional state of a given utterance by integrating contextual information from different modalities, which often includes subtle personal emotional states such as happiness, anger, and hatred \cite{poria2019emotionrecognitionconversationresearch, gong-etal-2024-mapping}, thereby improving the effectiveness of emotion recognition in dialogues.
Figure \ref{tag:introduction} illustrates an example of ERC with textual, acoustic, and visual inputs.

Multimodal emotion recognition (MER) itself has gained increasing attention due to the challenges of integrating diverse modalities, prompting research in both non-conversational and conversational settings. Existing surveys such as \citet{geetha2024multimodal} and \citet{gladys2023survey} focus on non-conversational MER but overlook key aspects like interlocutor modeling and context. \citet{fu2023emotion} reviews both unimodal and multimodal conversational MER, yet primarily centers on feature fusion, offering limited insight into core challenges such as cross-modal alignment, reasoning, modality missingness, and conflicts.

Despite growing interest, the MERC task remains underexplored. Existing surveys \cite{fu2023emotion,zhang2024survey} also lag behind recent advances, particularly the rise of multimodal large language models (MLLMs). To bridge this gap, we present a timely and comprehensive review of MERC. 
We first introduce the task definition and our survey methodology (§\ref{tab:task_setting}), benchmark datasets and evaluation methods (§\ref{tab:Data and Eval}), followed by a review of preprocessing techniques (§\ref{tab:Feature Processing}); then categorize recent methods (§\ref{tab:Methodology}) and outline key challenges and prospects (§\ref{tab:Challenges and Prospects}). 

In summary, the specific contributions of this survey are threefold:
\begin{itemize}[leftmargin=*]
    \item \textbf{Compilation of recent MERC research developments.} We systematically review and integrate the latest research progress made by MERC in recent years, covering diverse datasets and methodologies. 
    \item \textbf{Summarizing and comparing various MERC methods.} We evaluate the strengths and limitations of various MERC approaches, offering theoretical insights and practical guidance to help researchers and practitioners select appropriate methods.
    \item \textbf{Proposing challenges and future directions.} We identify key open issues in the MERC domain and put forward several potential future research directions, aiming to guide ongoing and future investigations by researchers and practitioners in MERC.
\end{itemize}

\section{Task Settings and Review Methodology}
\label{tab:task_setting}
In this section, we present the task settings of MERC and outline the methodology employed to compile the content of this survey, which details the strategy and selection criteria used to curate the final content for this survey paper.

\paragraph{Modalities.}
In the context of MERC, we define a modality as a distinct source or channel of information that conveys emotional or communicative signals. These modalities correspond to observable representations derived from different sensory inputs and are typically processed in separate feature spaces. The most common modalities include:

\begin{itemize}[leftmargin=*]
\item \textbf{Textual modality}: The transcribed representation of spoken utterances, capturing the semantic and syntactic content of language.

\item \textbf{Acoustic modality}: Prosodic and paralinguistic features extracted from speech, such as tone, pitch, and energy.

\item \textbf{Visual modality}: Non-verbal cues such as facial expressions, head movements, eye gaze, and gestures.
\end{itemize}

\paragraph{Task Definition.}
Given a dialogue $D = \{u_1, u_2, \ldots, u_N\}$ consisting of $N$ utterances spoken by multiple speakers, the goal of the MERC task is to predict an emotion label $e_i \in \mathcal{Y}$ for each utterance $u_i$. Each utterance is associated with three modalities: textual ($u_i^t$), acoustic ($u_i^a$), and visual ($u_i^v$), which provide complementary information for emotion recognition.
The multimodal representation of an utterance is denoted as:
\begin{equation}
u_i = [u_i^t; u_i^a; u_i^v], \quad \text{for } i = 1, 2, \ldots, N
\end{equation}
where $[\,\cdot\,; \cdot\,; \cdot\,]$ denotes combination of modalities.

\paragraph{Methodology for Literature Compilation:}
\paragraph{}
\textbf{\textit{Strategy.}} We conduct a comprehensive literature search using sources such as the ACL Anthology, Google Scholar, and general search engines (e.g., Google Chrome). Within the ACL Anthology, we focus on top venues including EMNLP, ACL, NAACL, and related workshops.\footnote{We used keywords such as \textit{``multimodal emotion recognition'', ``emotion recognition in conversation'', ``multimodal affective computing'', ``dialogue emotion recognition'', ``MERC benchmark'', ``context-aware emotion detection'',} and \textit{``multimodal conversational modeling''}.}
    
\textbf{\textit{Selection Criteria.}} We select papers that are directly relevant to MERC, with a focus on works that used at least two modalities (e.g., text, audio, visual), included conversational context, and evaluated on benchmark datasets such as IEMOCAP, MELD, and CMU-MOSEI. We prioritize recent papers from 2020 onward to reflect the state-of-the-art, while including foundational work when appropriate for historical context. The selection is based on a careful review of the
abstract, introduction, conclusion, and limitations of each paper.

\section{Datasets and Evaluation}
\label{tab:Data and Eval}
In this section, we describe the evaluation datasets and the evaluation metrics used for the MERC task, focusing on multimodal resources across multiple languages. For more detailed information about the single benchmarks, see Appendix §\ref{tab:dataset_detail}.

\begin{itemize}
    \item[(1)] \textit{English-centered:} IEMOCAP, MELD, CMU-MOSEI, AVEC, EmoryNLP, and MEmoR dataset.
    \item[(2)] \textit{Non-English:} M-MELD (French, Spanish, Greek, Polish), ACE (African), M\textsuperscript{3}ED (Mandarin).
\end{itemize}

As shown in Table \ref{tab:dataset_overview}, the domains covered by multimodal datasets have become increasingly diverse over time. The sources of these datasets include TV series, videos, and movies. At the same time, linguistic diversity has expanded to include languages such as French, Spanish, Greek, Polish, and Mandarin. Notably, there has also been a growing emergence of datasets targeting low-resource languages, such as African languages.
\begin{table}[htbp]
\centering
\renewcommand\arraystretch{1.1}
\scriptsize
\setlength\tabcolsep{2pt}
\begin{tabular}{lccr}
\toprule
\textbf{Datasets} & \textbf{Lang.}  & \textbf{Source}& \textbf{Year} \\
\midrule
IEMOCAP \cite{Busso_Bulut_Lee_Kazemzadeh_Mower_Kim_Chang_Lee_Narayanan_2008}& en & Videos & 2008 \\
AVEC \cite{DBLP:conf/icmi/SchullerVCP12}& en & Videos & 2012 \\
EmoryNLP \cite{DBLP:journals/corr/abs-1708-04299}& en &  TV series & 2017 \\
CMU-MOSEI \cite{Bagher_Zadeh_Liang_Poria_Cambria_Morency_2018}& en  & Videos&2018 \\
MELD \cite{Poria_Hazarika_Majumder_Naik_Cambria_Mihalcea_2019}& en &  TV series &2019\\
MEmoR \cite{shen2020memor}& en & Videos & 2020 \\
M\textsuperscript{3}ED \cite{zhao-etal-2022-m3ed}& zh & TV series &2022  \\
M-MELD \cite{Ghosh2022MMELDAM}& fr,es,el,pl & TV series &2023 \\
ACE \cite{sasu2025akan}& Akan & Movies &2025 \\
\bottomrule
\end{tabular}
\caption{Overview of Datasets.}
\label{tab:dataset_overview}
\end{table}
\paragraph{Datasets.} We divide the existing mainstream dataset into the following two categories: 

\paragraph{Evaluation Metrics.} Existing studies \citep[][\emph{inter alia}]{Majumder2019,ghosal-etal-2019-dialoguegcn} typically adopt multiple evaluation metrics to comprehensively assess the overall performance of models, including \textbf{Accuracy} \citep[e.g.,][]{shou2024efficient}, \textbf{Weighted-F1} \citep[e.g.,][]{ma2023transformer}, \textbf{Macro-F1} \citep[e.g.,][]{chudasama2022m2fnet}, and \textbf{Micro-F1} \citep[e.g.,][]{xie-etal-2021-knowledge-interactive} scores. To enable fine-grained analysis, these works also report per-emotion metric scores.

\section{Feature Processing}
\label{tab:Feature Processing}
Preprocessing dataset features is essential for effectively extracting meaningful information. We summarize feature preprocessing methods used in prior MERC research and analyze the typical preprocessing pipeline, which is often tailored to conversational settings. Specifically, we distinguish between two key components: \textbf{feature extraction} and \textbf{context modeling}.

\subsection{Feature Extraction}
For effective multimodal analysis, features must first be extracted from each modality stream (text, audio, visual). Mainstream approaches \citep[e.g.,][]{shi2023multiemo} typically process these modalities separately at this initial stage. While the core extraction techniques often overlap, the key difference in the multimodal setting lies in the objective and subsequent use of these features. In unimodal ERC, the extractor aims to capture enough information within that single modality for emotion classification. Table \ref{tab:feature_extraction_models} provides an overview of common feature extraction models employed in the multimodal studies surveyed in this paper.

\begin{table}[htbp]
\centering
\renewcommand\arraystretch{0.9}
\scriptsize
\setlength\tabcolsep{9pt}
\begin{tabular}{l|c}
\toprule
\textbf{Modality} & \textbf{Models} \\
\midrule
&LSTM \cite{Hochreiter_Schmidhuber_1997} \\
 & CNN \cite{Kim_2014_LSTM,Tran_Bourdev_Fergus_Torresani_Paluri_2015} \\
\textbf{Text} & Transformer \cite{vaswani2017attention}\\
&RoBERTa \cite{Liu2019RoBERTaAR}\\

&sBERT \cite{Reimers2019SentenceBERTSE}\\
\midrule
 & 3D-CNN \cite{Tran_Bourdev_Fergus_Torresani_Paluri_2015} \\
& OpenFace \cite{7477553}\\
\textbf{Visual}&MTCNN \cite{zhang2016joint} \\
&DenseNet \cite{Huang_Liu_Van_Der_Maaten_Weinberger_2017}\\
&VisExtNet \cite{shi2023multiemo}\\

\midrule
 \multirow{4}{*}{\textbf{Audio}}& openSMILE \cite{10.1145/1873951.1874246}  \\
 &COVAREP \cite{Degottex2014COVAREPA}\\
& librosa \cite{McFee2015librosaAA}\\
&DialogueRNN \cite{Majumder2019}\\
\bottomrule
\end{tabular}
\caption{Feature Extraction models.}
\label{tab:feature_extraction_models}
\end{table}

\begin{figure*}[htbp]
\scriptsize
\tikzset{
    basic/.style  = {draw, text width=1.45cm, align=center, 
    rectangle, fill=green!10},
    root/.style   = {basic, rounded corners=2pt, thin, align=center, fill=gray!10, rotate=0},
    tnode/.style = {basic, thin, rounded corners=2pt, align=left, fill=pink!30, text width=6.7cm, anchor=center},
    xnode/.style = {basic, thin, rounded corners=2pt, align=center, fill=blue!10,text width=3cm, anchor=center}, 
    ynode/.style = {basic, thin, rounded corners=2pt, align=center, text width=3.5cm, fill=yellow!10,anchor=center}, 
    edge from parent/.style={draw=black, edge from parent fork right}
}
\begin{forest} 
for tree={
    grow=east,
    growth parent anchor=east,
    parent anchor=east,
    child anchor=west,
    edge path={\noexpand\path[\forestoption{edge}, ->, >={latex}] 
        (!u.parent anchor) -- +(10pt,0) |- (.child anchor) \forestoption{edge label};},
}
    [Methods (\S\ref{tab:Methodology}), root,  text width=1cm, l sep=7mm,
        [Generation-based (\S\ref{tab:Generation-based Methods}), xnode, text width=1.8cm,l sep=7mm,
            [Lightweight Multimodal Fusion and Adaptation, ynode, l sep=7mm,
                [E.g. UniMSE \cite{hu2022unimse}; UniSA \cite{li2023unisa}; FacingBot \cite{tanioka2024toward}; MSE-Adapter \cite{yang2025mse}, tnode]],
            [Behavior-Aware and Multi-modal Instruction-Tuned, ynode, l sep=7mm,
                [E.g. DialogueLLM \cite{zhang2024dialoguellm}; BeMERC \cite{fu2025bemerc}; MERITS-L \cite{dutta2025llmsupervisedpretrainingmultimodal}, tnode]],
            [Instruction-Tuned with Speaker and Context Modeling, ynode, l sep=7mm,
                [E.g. InstructERC \cite{lei2023instructerc}; CKERC \cite{fu2024ckerc}; LaERC-S \cite{fu2025laerc}, tnode]]],
        [Fusion-based (\S\ref{tab:Fusion-based Methods}), xnode, text width=1.8cm,l sep=7mm,
            [Text-Dominant Modality, ynode, l sep=7mm,
                [E.g. MMT \cite{zou2022improving}; MPT \cite{zou2023multimodal}; MM-NodeFormer \cite{huang2024mm}; CMATH \cite{zhu2024cmath}, tnode]],
            [Equal Modality Weights, ynode, l sep=7mm, 
                [E.g. DialogueTRM \cite{mao2020dialoguetrm}; Emoformer \cite{li2022emocaps}; CMCF-SRNet \cite{zhang2023cross}; SDT \cite{ma2023transformer}, tnode]]],
        [Graph-based (\S\ref{tab:Graph-based Methods}), xnode, text width=1.8cm,l sep=7mm,
            [Fourier Graph Neural Networks, ynode, l sep=7mm,
                [E.g. ADGCN \cite{liu2023adaptive}; DGODE \cite{shou2025dynamic}; TL-RGCN \cite{su5011893multimodal}; GS-MCC \cite{ai2025revisiting}, tnode]],
            [Hypergraph Neural Networks, ynode, l sep=7mm,
                [E.g. GCNet \cite{lian2023gcnet}; MER-HGraph \cite{li2023multimodal}; MDH \cite{huang2024dynamic}; ConxGNN \cite{van2025effective}, tnode]],
            [Traditional Graph Neural Networks, ynode, l sep=7mm,
                [E.g. DialogueGCN \cite{ghosal-etal-2019-dialoguegcn}; MMGCN \cite{hu-etal-2021-mmgcn}; CORECT \cite{nguyen2023conversation}; GraphSmile \cite{li2024tracing}; MERC-GCN \cite{feng2025cross}, tnode]]]
    ]
\end{forest}
\caption{A taxonomy of mainstream methods.}
\label{fig:taxonomy_methods}
\end{figure*}
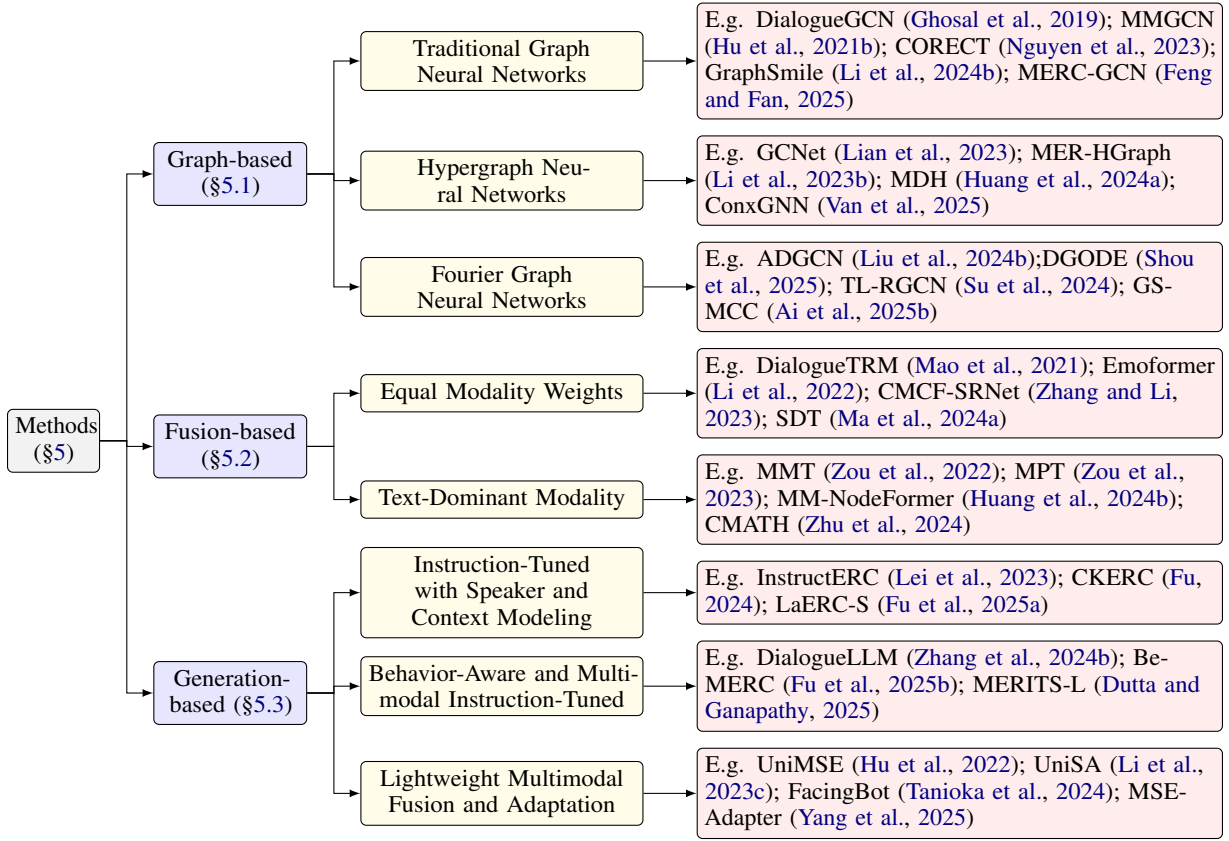

\subsection{Context Modeling} Context modeling primarily involves two types of contextual dependencies: \textbf{situation-level} and \textbf{speaker-level} modeling.

\paragraph{Situation-level.} The emotional state of a speaker is influenced not only by the semantic content of the current utterance but also by the surrounding contextual semantics. Therefore, existing approaches \cite{hu-etal-2021-dialoguecrn,Majumder2019} commonly employ specialized networks to model the sequential dependencies among utterances, aiming to more accurately capture the speaker's temporal emotional dynamics. Given the textual feature $u_i \in \mathbb{R}^{d_u}$ for each utterance, the sequential context representation $c^s_i \in \mathbb{R}^{2d_u}$ is computed as follows:
\begin{equation}
\mathbf{c}_{i}^{s},\ \mathbf{h}_{i}^{s} = {\mathrm{Model}}\left(\mathbf{u}_{i},\ \mathbf{h}_{i-1}^{s}\right)
\end{equation}
where $h^s_i \in \mathbb{R}^{d_u}$ represents the $i$-th hidden state of the contextual sequence.

    
\paragraph{Speaker-level.} Speaker identity information typically exhibits temporal and relational properties of emotions, which can enhance the model's ability to perceive speaker role information. Therefore, to more effectively learn and distinguish speaker-level contextual representations, many studies have further introduced speaker-related structured information on top of dialogue context modeling. Common approaches include using \emph{\textbf{Speaker Embeddings}} \cite{ma2023transformer,shen2020memor} to explicitly differentiate between different speakers, or leveraging \emph{\textbf{Graph Neural Networks}} \cite{ai2024gcn,van2025effective} to construct interaction graphs between speakers, thereby more comprehensively modeling the dependencies between them:


\emph{\textbf{Speaker Embeddings}}: The speaker embedding $S_i$ is combined with the modality features (e.g., text, audio, or vision) to produce modality representations that are speaker- and context-aware.

\begin{equation}
\mathbf{x}_m = \mathbf{c}_{i}^{s} + \mathbf{S}_i, \quad m \in \{t, a, v\}
\end{equation}


\emph{\textbf{Graph Neural Networks}}:
Speaker interactions can be further modeled by constructing a dialogue graph to capture inter-utterance and inter-speaker dependencies beyond sequential semantics. A dialogue graph is typically defined as $G = (V, E, W, R)$, where each node $v_i \in V$ corresponds to the $i$-th utterance, and its associated feature vector $\mathbf{c}_i^s$ is obtained from sequential modeling of contextual dependencies. Edges $e_{ij} \in E$ represent interaction links between utterances, with associated weights $\omega_{ij} \in W$ reflecting the interaction strength and types $r_{ij} \in R$ encoding speaker-related or structural relationships.

Based on this graph, the context-aware representation $\mathbf{h}_i^g$ for node $v_i$ is computed using a graph neural network as follows:

\begin{equation}
\mathbf{h}_i^{g} = \mathrm{GNN}(\mathbf{c}_i^s,\ \{ (\mathbf{c}_j^s, \omega_{ij}, r_{ij}) \mid e_{ij} \in E \})
\end{equation}
where $\mathbf{c}_j^s$ are the contextual node features from neighboring utterances. The GNN aggregates information from connected nodes to enhance $\mathbf{c}_i^s$ with speaker- and structure-aware interaction knowledge.


\section{Methodology}
\label{tab:Methodology}
This section discusses state-of-the-art approaches to the MERC tasks.
We summarize them from three perspectives: Graph-based Methods (§\ref{tab:Graph-based Methods}), Fusion-based Methods (§\ref{tab:Fusion-based Methods}), and Generation-based Methods (§\ref{tab:Generation-based Methods}). An overview of the methods and subcategories with representative examples is presented in Figure \ref{fig:taxonomy_methods}. 

It is worth noticing that some methods in MERC inherently involve multiple components (e.g., fusion modules within graph-based or generation-based frameworks). Our taxonomy is not intended to be strictly disjoint, but rather to organize the literature based on the core modeling paradigm or innovation focus of each method. The categorization of methods is thus as follows:

\begin{itemize}[leftmargin=*]

\item \textbf{Graph-based}: Methods are categorized as graph-based when the primary architecture centers around graph neural networks for modeling conversational structure, even if auxiliary modules (e.g., fusion layers) are integrated.

\item \textbf{Fusion-based}: Methods are grouped under fusion-based when their main contribution lies in the design of cross-modal interaction mechanisms, regardless of the backbone architecture (e.g., Transformer, LSTM).

\item \textbf{Generation-based}: Generation-based methods refer to recent approaches that leverage LLMs to generate predictions or intermediate reasoning, often using prompt engineering or instruction tuning, even if lightweight fusion components are present.

\end{itemize}

\subsection{Graph-based Methods}
\label{tab:Graph-based Methods}
\begin{figure}[htbp]
    \centering
    \includegraphics[width=1\linewidth]{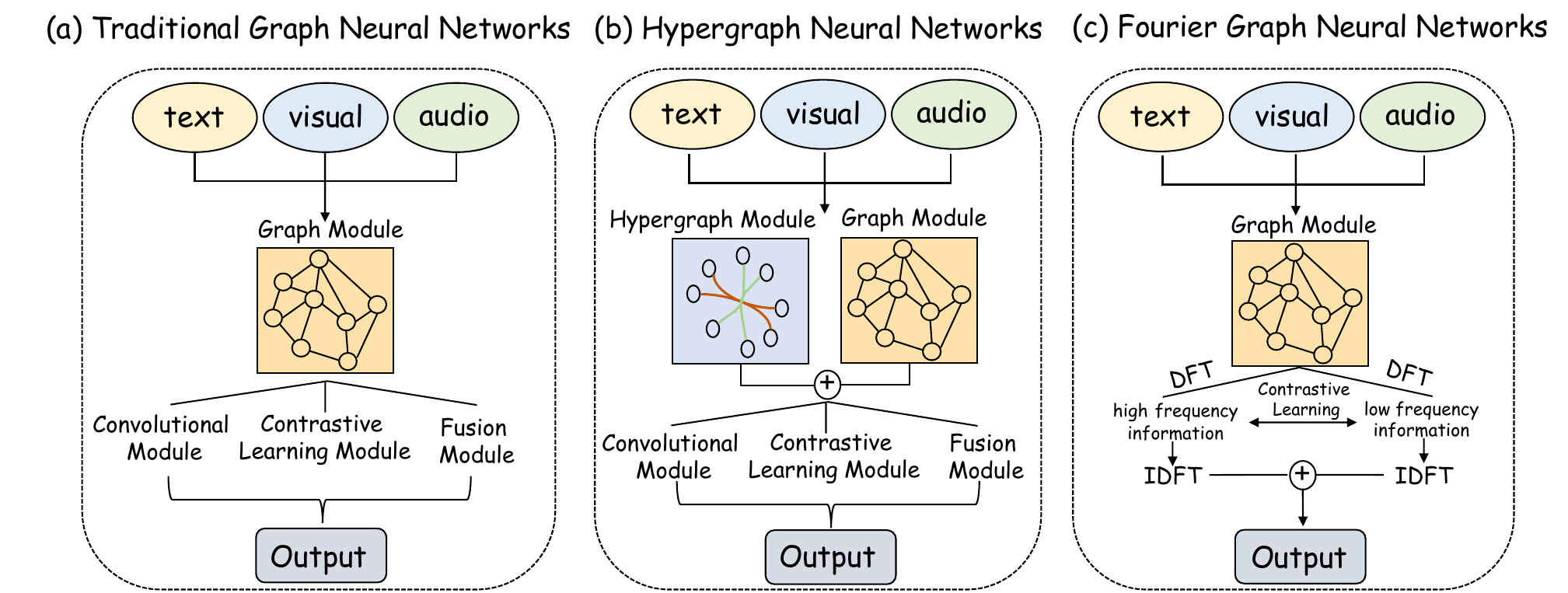}
    \caption{Development of Graph-based Methods.}
    \label{fig:graph_method}
\end{figure}

Dialogues can be naturally interpreted as graph structures due to the intrinsic correlations and dependencies among utterances. Conversations often feature multi-turn interactions with complex dependency and interaction patterns, which can be effectively modeled through the edge structures of Graph Neural Networks (GNNs) \cite {scarselli2008graph}. With the increasing interest in multimodal dialogue understanding, GNNs have evolved beyond their application \cite{liu2024graph} in textual data to embrace multimodal inputs. Besides, recent methods also integrate auxiliary modules (e.g., convolution, contrastive learning, and fusion) to enhance the performance.
Figure \ref{fig:graph_method} illustrates recent advancements in graph-based methods. We categorize them into traditional graphs, hypergraphs, and fourier graph neural networks.

\textbf{Traditional Graph Neural Networks.} 
Previous works such as bc-LSTM \cite{poria-etal-2017-context} and ICON \cite{hazarika2018conversational} primarily relied on sequential approaches. DialogueGCN \cite{ghosal-etal-2019-dialoguegcn} was the first to introduce GNNs into ERC, addressing the limitations of earlier sequence-based models like DialogueRNN \cite{Majumder2019} in capturing contextual dependencies. To effectively integrate information from different modalities, \citet{hu-etal-2021-mmgcn} constructed a graph structure that fuses multimodal features, enabling the model to capture inter-modal dependencies through graph convolutional networks and incorporating speaker information to enhance the representation of conversational semantics. Inspired by the application of graph convolutions in ERC, \citet{li2024tracing} proposed the GSF module, which introduces an alternating graph convolution mechanism to hierarchically extract both cross-modal and intra-modal emotional information. Some studies further enhanced graph-based models with attention mechanisms for multimodal fusion; for example, \citet{feng2025cross} integrated a Cross-modal Attention Module to better fuse information from different modalities, while \citet{nguyen2023conversation} designed a cross-modal attention mechanism to explicitly model the heterogeneity between modalities.

\textbf{Hypergraph Neural Networks.} Although traditional graph-based methods can capture long-range and multimodal contextual information, they are often challenged by missing modalities during conversations. \citet{lian2023gcnet} tackled this issue by jointly optimizing classification and reconstruction tasks in an end-to-end manner to effectively model incomplete data. The related works \cite{li2023multimodal,huang2024dynamic} considered the limitations imposed by the pairwise relationships between GNNs nodes.\citet{van2025effective} constructed a multimodal fusion graph and introduced Hypergraph Neural Networks \cite{feng2019hypergraph} to connect multiple modalities or utterance nodes simultaneously, thereby capturing more complex multivariate dependencies and high-order interactions in conversations, and enhancing the modeling of emotion propagation.

\textbf{Fourier Graph Neural Networks.} Increasing the depth of GNN layers can lead to the over-smoothing problem \cite{liu2022revisiting, yi2023fouriergnn}, which hampers the modeling of long-range semantic dependencies and complementary modality relations. To address this issue, GS-MMC \cite{ai2025revisiting} proposes a graph-based framework for multimodal consistency and complementarity learning. This method employs a Fourier graph operator to extract high- and low-frequency emotional signals from the frequency domain, capturing both local variations and global semantic trends. Additionally, a contrastive learning mechanism \cite{oord2018representation} is designed to enhance the semantic consistency and complementarity of these signals in a self-supervised manner, thereby improving the model's ability to recognize true emotional states.

Graph-based methods effectively capture long-range dependencies and speaker interactions by modeling utterances as nodes and relations as edges. In traditional NLP tasks, they are particularly effective for structured inputs such as token-level representations \cite{Zhang2023token,Zhang2024What}. 
However, integrating heterogeneous multimodal signals into graph structures remains challenging, as naïve connections may introduce noise without proper modality alignment.

\subsection{Fusion-based Methods}
\label{tab:Fusion-based Methods}

\begin{figure}[htbp]
    \centering
    \includegraphics[width=0.8\linewidth]{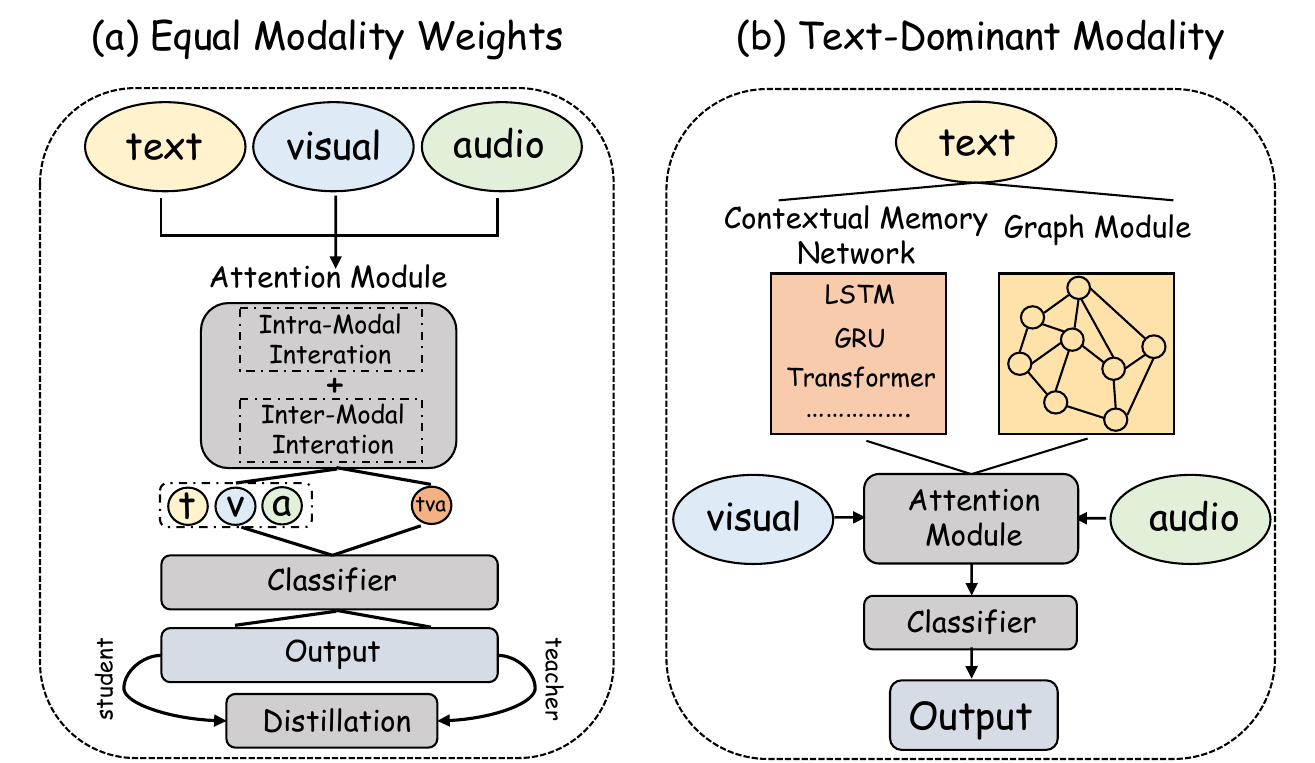}
    \caption{Development of Fusion-based Methods.}
    \label{fig:fusion_method}
\end{figure}

In MERC, effective fusion of heterogeneous multimodal features is crucial but challenging due to noise introduced during interaction modeling. The advancements of the Transformers architecture \cite{vaswani2017attention}, with its self-attention mechanism, promote advancements of the MERC methods for capturing cross-modal and contextual dependencies. 
To enhance cross-modal interactions, recent methods build on Transformers with tailored fusion strategies. We refer to these methods as fusion-based and illustrate them in Figure \ref{fig:fusion_method}. Some approaches promote equal interaction among modalities to improve robustness, while others adopt a primary-auxiliary scheme, typically using text as the core, with other modalities providing complementary signals.

\textbf{Equal Modality Weights.} Equal interaction can fully utilize information from various modalities, preventing over-reliance on a single modality. \citet{li2022emocaps} proposed achieving emotion recognition by equally integrating emotional vectors and sentence vectors from different modalities to form emotion capsules. \citet{zhang2023cross} designed a local constraint module for modalities within the Transformer to promote modality interaction and incorporates a semantic graph to address the lack of semantic relationship information between utterances. \citet{mao2020dialoguetrm} constructed a hierarchical Transformer where each modality can flexibly switch between sequential and feedforward structures based on contextual information. Inspired by hierarchical modality interaction, \citet{ma2023transformer} introduced a hierarchical gating \cite{ma2019hierarchical} fusion strategy into the Transformer architecture to enable fine-grained modality interaction and designs self-distillation \cite{zhang2019your} to further learn better modality representations.

\textbf{Text-Dominant Modality.} 
Some methods propose models based on primary–auxiliary modality collaboration, where auxiliary modalities are used to enhance the performance of the primary (textual) modality. 
\citet{huang2024mm} suggested that enhancing a text-dominant model with auxiliary modalities can improve performance. \citet{zou2022improving} employed a Transformer architecture to design cross-modal attention for learning fusion relationships between different modalities, preserving the integrity of the primary modality's features while enhancing the representation of weaker modality features. It also uses a two-stage emotional cue extractor to extract emotional evidence. Building on this, \citet{zou2023multimodal} proposed using weaker modalities as multimodal prompts while performing deep emotional cue extraction for stronger modalities. The cue information is embedded into various attention layers of the Transformer to facilitate the fusion of information between the primary and auxiliary modalities. \citet{zhu2024cmath} introduced an asymmetric CMA-Transformer module for central and auxiliary modalities to obtain fused modality information and proposes a hierarchical distillation \cite{yang2021hierarchical} framework to perform coarse- and fine-grained distillation. This approach ensures the consistency of modality fusion information at different granularities.

Fusion-based methods focus on learning cross-modal interactions through attention mechanisms, with Transformer-based models achieving strong generalization. These approaches are efficient for tasks with well-aligned modality inputs but often overlook dialogue-level structures such as speaker dependencies. Compared to graph-based models, they emphasize modality-level fusion over relational reasoning.

\subsection{Generation-based Methods}
\label{tab:Generation-based Methods}
\begin{figure}[htbp]
    \centering
    \includegraphics[width=1\linewidth]{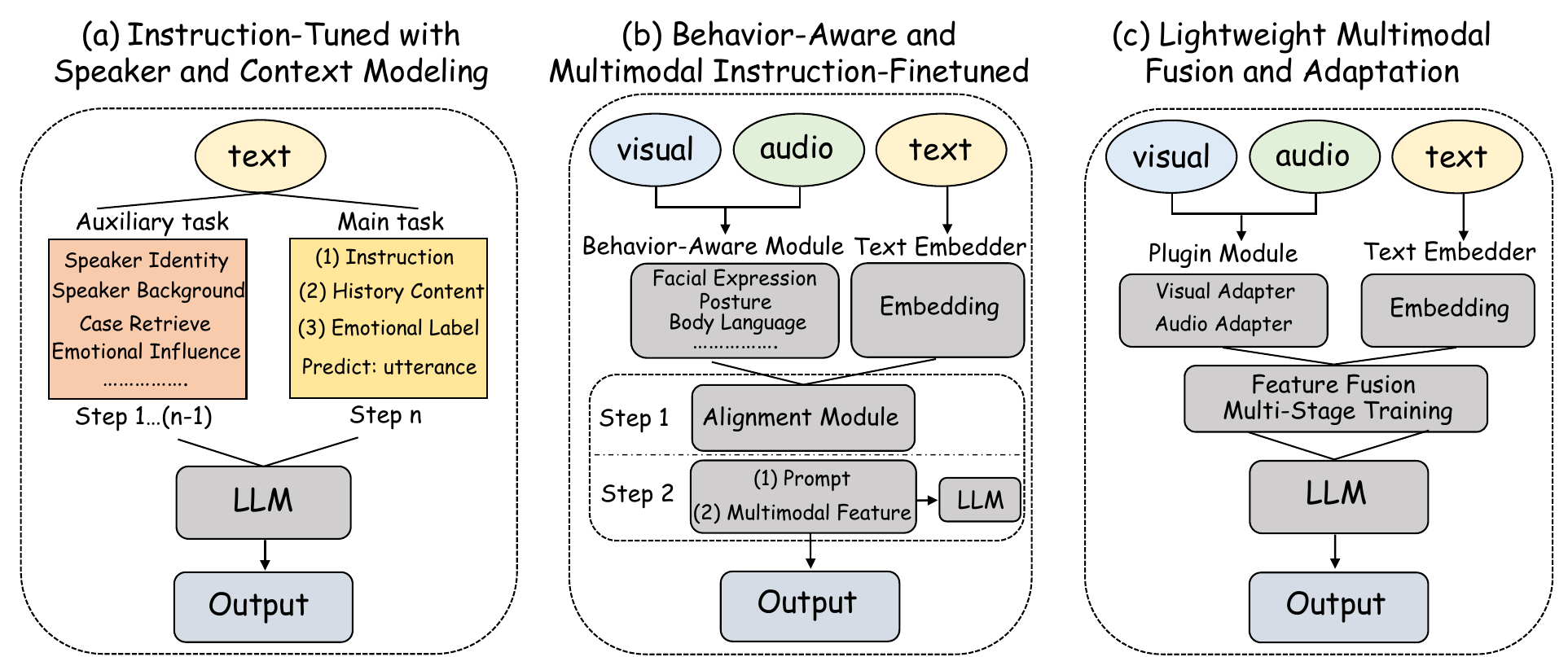}
    \caption{Development of Generation-based Methods.}
    \label{fig:generation_method}
\end{figure}

In recent years, pretrained LLMs have achieved remarkable success in natural language processing tasks \cite{chu2024qwen2}, demonstrating strong emergent capabilities \cite{wei2022chain}. However, despite their powerful general-purpose abilities, leveraging their full potential in specific sub-tasks still requires carefully crafted, high-quality prompts \cite{wei2021finetuned} to bridge the gap in reasoning capabilities.As shown in Figure \ref{fig:generation_method}, researchers have proposed various model improvement strategies to effectively integrate contextual and multimodal information into LLMs while addressing their substantial computational resource demands. 

\textbf{Instruction-Tuned with Speaker and Context Modeling.} 
ERC tasks have predominantly relied on discriminative modeling frameworks. With the emergence of LLMs, InstructERC \cite{lei2023instructerc} was the first to propose a generative framework for ERC. It introduces a simple yet effective retrieval-based prompting module that helps LLMs explicitly integrate multi-granularity supervisory signals from dialogues. Additionally, it incorporates an auxiliary emotion alignment task to better model the complex emotional transitions between interlocutors in conversations. Inspired by the integration of commonsense knowledge in COSMIC \cite{ghosal2020cosmic},  recent work \cite{fu2024ckerc,fu2025laerc} designed a prompt generation approach based on dialogue history to elicit speaker-related commonsense using LLMs by injecting commonsense knowledge into ERC.

\textbf{Behavior-Aware and Multimodal Instruction-Tuned.} 
To address the lack of multimodal integration, \citet{dutta2025llmsupervisedpretrainingmultimodal} incorporated both acoustic and textual modalities. Considering that visual information may provide richer emotional cues, \citet{zhang2024dialoguellm} constructed a high-quality instruction dataset using image and text data, and fine-tunes the model using Low-Rank Adaptation (LoRA) \cite{song2024low}. Furthermore, \citet{fu2025bemerc} introduced a novel behavior-aware Multimodal LLM (MLLM)-based ERC framework. It consists of three core components: a video-derived behavior generation module, a behavior alignment and refinement module, and an instruction tuning module \cite{wei2021finetuned}. The first two modules enable the model to infer human behaviors from limited information, thereby enhancing its behavioral perception capability. The instruction tuning module improves the model’s emotion recognition performance by aligning and fine-tuning the concatenated multimodal inputs.

\textbf{Lightweight Multimodal Fusion and Adaptation.} As LLMs become increasingly large, the computational costs for ERC also rise significantly \cite{zhang2024dialoguellm}. Inspired by domain-specific LLM paradigms tailored for affective computing \cite{hu2022unimse, li2023unisa,tanioka2024toward}, MSE-Adapter \cite{yang2025mse} proposed a lightweight and adaptable plug-in architecture with two modules: TGM, for aligning textual and non-textual features, and MSF, for multi-scale cross-modal fusion. Built on a frozen LLM backbone and trained via backpropagation, it enables efficient and multimodally-aware ERC with minimal computational cost. 
Similarly, SpeechCueLLM \cite{wu-etal-2025-beyond} introduced a lightweight plug-in that converts speech features into natural language prompts, enabling LLMs to perform multimodal emotion recognition without architectural changes.

Generation-based methods leverage LLMs to reformulate ERC as a text generation task, enabling flexible adaptation across datasets and domains. Their end-to-end nature simplifies input processing but limits fine-grained control over multimodal integration. In contrast to structured models, LLMs excel at scalability but require further refinement to model multimodal dependencies explicitly.

\section{Challenges and Prospects}
\label{tab:Challenges and Prospects}

Based on current trends and developments in the MERC task, in this section, we outline several existing challenges and open questions, highlighting opportunities for future improvements. We structure our discussion along a logical progression: starting with foundational limitations in data collection and FAIR compliance, we then examine challenges in multimodal modeling and conclude with considerations for real-world deployment. This trajectory reflects how upstream issues in data and modeling propagate downstream, ultimately shaping the robustness, inclusivity, and applicability of MERC systems.

\textbf{FAIR-related Issues Pose Challenges in MERC.} 
The FAIR principles provide guidelines for improving the Findability, Accessibility, Interoperability, and Reusability of digital assets \cite{wilkinson2016fair}. Collecting large and diverse multimodal emotion data is costly and time-consuming; some large dialogue datasets (e.g., M\textsuperscript{3}ED, \citealp{zhao-etal-2022-m3ed}) are still monolingual and domain-bound. These limitations directly conflict with the FAIR principles. Some ERC datasets lack rich metadata or persistent identifiers, undermining findability and interoperability. Others are subject to access restrictions or copyright constraints, while many adopt inconsistent labeling schemes that hinder reusability. Consequently, researchers often have to train on small or biased samples, which undermines generalization and the reuse of models. To address these issues, future work could prioritize the development of multilingual benchmark datasets with standardized metadata and open licensing, possibly through collaborative consortiums that align with FAIR principles.

\textbf{Low-Resource, Multilingual, and Multicultural Settings.} 
As described in the previous paragraph, most state-of-the-art MERC systems are trained on English-language datasets, which limits their global applicability. Although building a large-scale, diverse MER corpus is essential, it remains an obvious challenge as it requires expert annotation of data. The limited annotated data forces researchers to rely on transfer learning \cite{ananthram2020multi}, zero-shot \cite{qi2021zero}, or few-shot methods \cite{yang2023few}. However, data scarcity and the high cost of emotion annotation continue to be major obstacles for MER in low-resource domains \cite{hussain2025low}. Emotions are expressed differently across languages and cultures, further compounding the challenges of MER. Variations in emotional expression and interpretation due to cultural differences can lead to inconsistencies in labeling. Most existing corpora are culture-specific, limiting their generalizability. Although researchers have acknowledged this challenge \cite{vetriselvi2024sentiment, devi2024multi}, MER systems aiming for global applicability must account for both linguistic diversity and culturally driven display rules. Future work could explore culture-adaptive pretraining and cross-lingual transfer learning methods that embed culturally sensitive emotion semantics across languages.

\textbf{Complexities of Fusion Strategies across Modalities.}
Multimodal fusion techniques include early fusion, mid-level fusion, late fusion, hybrid fusion, and others \cite{atrey2010multimodal,gandhi2023multimodal}. A key challenge is that conversational signals, such as voice, facial expressions, and transcripts, are inherently asynchronous and occur at different time scales, making it difficult to align them at the utterance level. Emotions also depend on the context of preceding and subsequent conversation turns, so the model must capture temporal dynamics \cite{wang2024dynamic}. Previous studies have used recurrent or self-attention layers to model sequential context \cite{houssein2024tfcnn, dutta2024deep}, but long-range dependencies remain challenging to learn. How to balance and integrate contextual sentiment cue features with multimodal fusion features in decision-making, and how to determine which fusion strategies are most effective across different modalities, remain open and important research questions \cite{geetha2024multimodal, Ramaswamy}. Recent advances in adaptive fusion strategies and dynamic attention mechanisms show promise, and future methods could explore transformer-based fusion that dynamically reweights modalities based on conversational context.

\textbf{Cross-Modal Alignment, Noise Modality, Missing Modality, and Modality Conflicts.} 
Misaligned or inconsistent features can inhibit the ability of a model to fully utilize multimodal signals, affecting its robustness and generalization \cite{ma2024learning, li2024multimodal}. Noise modality, missing modality, or imbalanced modality distributions may bias simple fusion strategies \cite{mai2023multimodal, zhang2024multimodal}. Even when all modalities are available, they may convey conflicting emotional signals, further complicating fusion and decision-making. Perceiving the uncertainty of different modalities for feature enhancement and resolving conflicts between modality features are important areas that need further exploration in MER research. Therefore, exploring cross-modal transfer and fusion to improve generalization in ERC has attracted the attention of more and more researchers \cite{fan2024cross, li2024cfn, feng2025cross}. Some ERC methods incorporate variants of cross-modal attention \cite{guo2024speaker, krishna2020multimodal}, graph-based fusion \cite{li2023graphmft, hu-etal-2021-mmgcn}, or mutual learning to align features during training and improve cross-domain performance \cite{lian2021ctnet}. Future research can further investigate robust training frameworks that include modality dropout, uncertainty-aware fusion, or reinforcement learning to selectively attend to trustworthy modalities.

\textbf{Effective Modality Selection.}
Multimodal learning refers to the integration of information from various heterogeneous sources and aims to effectively leverage data from diverse modalities \cite{he2024efficient, tsai2024text}. In multimodal representation learning, not all modalities contribute equally to the task. Some modalities may introduce noise and need to be removed, while others may not be essential for the task at hand but could be indispensable for other subtasks. Existing research proposes modality selection algorithms that identify the contribution of each modality \cite{marinov2022modselect, mai2023multimodal}. However, selecting the most appropriate subset of modalities for a task remains one of the key challenges in multimodal learning. An emerging direction is to integrate learnable modality gates or sparsity-inducing regularization into fusion models to automatically suppress uninformative modalities during training.

\textbf{Efficient Fine-tuning Approach Using Multimodal LLMs.}
Multimodal LLMs have brought major advances in enabling machines to learn across modalities. Some models are increasingly used in MERC, offering zero-shot or few-shot generalization across different modalities \cite{li2024improving, yang2024emollm, bo2025toward}. The use of LLMs in MERC opens up new possibilities for capturing deeper semantic and conversational cues beyond surface-level emotion signals. However, efficiently fine-tuning these models for emotion understanding still presents challenges, especially in low-resource and culturally diverse settings. Efficient adaptation of MLLMs to capture the nuance of emotions across diverse datasets, languages, or cross-cultural settings remains an open research frontier. Promising future directions include using adapter modules or low-rank fine-tuning techniques to adapt large models to specific emotion tasks with minimal data.

\textbf{MERC Application.}
With the growing use of interactive machine applications, MERC has become a critical research area. Applications in human-computer interaction \cite{ahmad2023systematic, moin2023emotion}, healthcare \cite{ayata2020emotion, islam2024enhanced}, education \cite{villegas2025multimodal, vani2025multimodal}, and virtual collaboration demand robust and adaptable emotion recognition technologies that function effectively in naturalistic and dynamic environments. \citet{yang2022multimodal} investigated MER in contexts affected by face occlusions, such as those introduced by surgical and fabric masks.
\citet{khan2024exploring} studied contactless techniques in MER, surveying a range of nonintrusive modalities (e.g., visual cues, physiological signals). \citet{huang2024real} developed a MER system for online learning to enable real-time monitoring and feedback on learners' emotional states. These studies highlight key directions for advancing MERC systems, particularly to make them more robust and context-aware. The ongoing research should continue to focus on real-world deployment scenarios. Future MERC systems could benefit from incremental learning techniques and user-in-the-loop feedback mechanisms that allow adaptation in dynamic, real-time environments.

\textbf{Expanding the Modality Space in Emotion Recognition.} 
Current MER systems mainly rely on vision, audio, and text, given their accessibility and prevalence in datasets. Yet human emotion is expressed through additional channels such as gaze, gestures, posture, and physiological signals (e.g., heart rate, skin conductance, brain activity) \cite{Noroozi2021,Udahemuka2024,wang2023,brainsci10100687,liu2024smilefacesadnesseyes}. These modalities remain underrepresented due to challenges in collecting synchronized, high-quality data and evolving annotation standards \cite{Udahemuka2024,brainsci10100687,KIM2024123723}. Expanding beyond the traditional three can reduce reliance on potentially misleading cues and open new application domains. For instance, bio-signals and contextual cues could enhance emotion sensing in health and education, while wearable sensors and eye-trackers may enable emotion-aware experiences in VR, driver monitoring, or social robotics. In summary, gaze, physiological, and other embodied modalities are promising but underexplored in affective computing. Future work should explore how to systematically integrate these diverse modalities into large-scale benchmarks and develop models capable of robustly leveraging them in real-world scenarios.

\section{Conclusion}

MERC seeks to understand emotions by integrating various modalities in to dialogue of linguistic, acoustic, visual signals, and beyond. While recent progress has introduced diverse modeling strategies, significant challenges remain in data scarcity, modality alignment, and generalization across languages and cultures.

This survey provides a structured review of the MERC landscape, compares representative approaches, and highlights key open research problems. We hope it serves as a practical reference to support the future development of robust and inclusive emotion recognition systems.


\section*{Limitations}
This survey offers a structured and up-to-date overview of MERC, with an emphasis on recent deep learning-based approaches. However, several limitations should be acknowledged to contextualize the scope of our work.

First, due to the focus on more advanced recent technologies, we provide only high-level summaries of representative methods, without delving into full technical details. Additionally, approaches developed prior to 2020 receive limited coverage, as our focus is primarily on recent trends that align with the rapid evolution of large-scale multimodal systems.

Second, our literature review is largely drawn from English-language publications in major conferences and repositories, including Interspeech, ICASSP, ACM, *ACL, ICML, AAAI, CVPR, COLING, and preprints on arXiv. While these venues represent core research communities in MERC, relevant contributions from other regions, languages, or domains may be underrepresented.

In the benchmark section, we highlight widely-used datasets, but do not aim for exhaustive comparison. For more in-depth benchmarking, we refer readers to complementary surveys such as \citet{zhao-etal-2022-m3ed}, \citet{sasu2025akan} and \citet{gan2024survey}.

Finally, while we identify several open challenges and underexplored directions, our discussion is not exhaustive. Rather than providing definitive answers, we aim to surface critical issues and foster further inquiry. We view these open questions as productive entry points for future research and hope this survey supports ongoing efforts to develop more advanced MERC systems.

\section*{Acknowledgments}
The authors acknowledge the use of ChatGPT exclusively to refine the text for grammatical checks in the final manuscript. This work was supported in part by the Guangdong Basic and Applied Basic Research Foundation under Grant 2023A1515011370, the National Natural Science Foundation of China (32371114), the Characteristic Innovation Projects of Guangdong Colleges and Universities (No. 2018KTSCX049), and the Guangdong Provincial Key Laboratory [No.2023B1212060076]. 
YL acknowledges support from the Duke Power Endowed Professorship at the School of Business, North Carolina Central University, and the IBM-HBCU Quantum Center Faculty Fellowship. 
ZG is supported by the National Science Foundation via ARNI (The NSF AI Institute for Artificial and Natural Intelligence), under the Columbia 2025 Research Project ("Towards Safe, Robust, Interpretable Dialogue Agents for Democratized Medical Care").

\bibliography{custom}

\appendix

\section{Datasets Details}
\label{tab:dataset_detail}
\paragraph{IEMOCAP.} 
\label{IEMOCAP}
The IEMOCAP dataset \cite{Busso_Bulut_Lee_Kazemzadeh_Mower_Kim_Chang_Lee_Narayanan_2008} consists of dyadic conversation videos from ten speakers, comprising 151 dialogues and 7,433 utterances. Sessions 1 to 4 are used as the training set, while the last session is held out as the test set. Each utterance is annotated with one of six emotion labels: happy, sad, neutral, angry, excited, and frustrated.
\paragraph{MELD.}
\label{MELD}
The Multimodal EmotionLines Dataset (MELD) \cite{Poria_Hazarika_Majumder_Naik_Cambria_Mihalcea_2019} is an extension of the EmotionLines corpus \cite{Chen_Hsu_Kuo_Ting-Hao_Huang_Ku_2018}, constructed from the TV series Friends. It contains 1,433 multi-party conversations and 13,708 utterances. Each utterance is annotated with one of seven emotion categories: anger, disgust, fear, joy, neutral, sadness, and surprise. Unlike dyadic datasets, MELD captures the complexity of multi-speaker interactions, making it well-suited for studying emotion recognition in multi-party conversational settings.
\paragraph{CMU-MOSEI.}
\label{CMU-MOSEI}
The CMU-MOSEI dataset \cite{Bagher_Zadeh_Liang_Poria_Cambria_Morency_2018} consists of 23,453 sentence-level video segments from over 1,000 speakers covering 250 topics, collected from YouTube monologue videos. Each segment is annotated for sentiment on a 7-point Likert scale ([-3: highly negative, -2: negative,
-1: weakly negative, 0: neutral, +1: weakly positive,
+2: positive, +3: highly positive]) and for the intensity of six Ekman \cite{ekman1980facial} emotions: happiness, sadness, anger, fear, disgust, and surprise. The dataset provides aligned language, visual, and acoustic modalities, making it a large-scale benchmark for multimodal sentiment and emotion recognition.

\paragraph{M\textsuperscript{3}ED.} The M\textsuperscript{3}ED dataset \cite{zhao-etal-2022-m3ed} consists of 990 dyadic dialogues and 24,449 utterances collected from 56 Chinese TV series. Each utterance is annotated with one or more of seven emotion labels: happy, surprise, sad, disgust, anger, fear, and neutral. The dataset covers text, audio, and visual modalities and features blended emotions and speaker metadata, making it the first large-scale multimodal emotional dialogue corpus in Chinese.

\paragraph{ACE.} The ACE dataset \cite{sasu2025akan} contains 385 emotion-labeled dialogues and 6,162 utterances in the Akan language, collected from 21 movie sources. It includes multimodal information across text, audio, and visual modalities, and is annotated with one of seven emotion categories: neutral, sadness, anger, fear, surprise, disgust, and happiness. Word-level prosodic prominence is also provided, making it the first such dataset for an African language. The dataset is gender-balanced, featuring 308 speakers, and is split into training, validation, and test sets in a 7:1.5:1.5 ratio.

\paragraph{MEmoR.} The MEmoR dataset \cite{shen2020memor} consists of 5,502 video clips and 8,536 person-level samples extracted from the sitcom The Big Bang Theory, annotated with 14 fine-grained emotions. Unlike most datasets focusing solely on speakers, MEmoR includes both speakers and non-speakers, even when modalities are partially or completely missing. Each sample comprises a target person, an emotion moment, and multimodal inputs (text, audio, visual, and personality features), making it a challenging benchmark for multimodal emotion reasoning beyond direct recognition.

\paragraph{AVEC.} The AVEC dataset \cite{DBLP:conf/icmi/SchullerVCP12}, derived from the SEMAINE corpus \cite{mckeown2011semaine}, features human-agent interactions annotated with four continuous affective dimensions: valence, arousal, expectancy, and power. While the original labels are provided at 0.2-second intervals, we aggregate them over each utterance to obtain utterance-level annotations for emotion analysis.

\paragraph{M-MELD.} M-MELD \cite{Ghosh2022MMELDAM} is a multilingual extension of the MELD dataset, created to support emotion recognition in conversations across different languages. While MELD is an English-only multimodal dataset, M-MELD includes human-translated versions of the original utterances in four additional languages: French, Spanish, Greek, and Polish. This multilingual corpus retains the original multimodal structure and emotion labels, enabling research in cross-lingual and multimodal emotion analysis. By balancing high-resource and low-resource languages, M-MELD offers a valuable benchmark for developing and evaluating multilingual ERC models.

\paragraph{EmoryNLP.} EmoryNLP \cite{DBLP:journals/corr/abs-1708-04299} is a textual emotion-labeled dataset derived from the Friends TV series, containing over 12,000 utterances across multi-party dialogues. Each utterance is annotated with one of seven emotions: neutral, joyful, peaceful, powerful, mad, sad, or scared, offering a fine-grained resource for emotion recognition in conversational settings.

\end{document}